\documentclass[11pt]{article}

\usepackage[preprint]{acl}

\usepackage{times}
\usepackage{latexsym}

\usepackage[T1]{fontenc}
\usepackage{amsmath} 
\usepackage[utf8]{inputenc}

\usepackage{microtype}

\usepackage{inconsolata}

\usepackage{graphicx}

\usepackage{booktabs}
\title{Aspect-Level Obfuscated Sentiment in Thai Financial Disclosures and Its Impact on Abnormal Returns}

\author{
 \textbf{Attapol T. Rutherford\textsuperscript{1}},
 \textbf{Sirisak Chueykamhang\textsuperscript{2}},\\
 \textbf{Thachaparn Bunditlurdruk\textsuperscript{1}}, 
 \textbf{Nanthicha Angsuwichitkul\textsuperscript{1}} \\
\\
\textsuperscript{1} Department of Linguistics, Faculty of Arts, Chulalongkorn University \\
\textsuperscript{2} Sasin Graduate Institute of Business Administration, Chulalongkorn University 
\\
\small \texttt{attapol.t@chula.ac.th, sirisak.chueykamhang@sasin.edu} 
}

\begin{document}
\maketitle

\begin{abstract}
Understanding sentiment in financial documents is crucial for gaining insights into market behavior. These reports often contain obfuscated language designed to present a positive or neutral outlook, even when underlying conditions may be less favorable. This paper presents a novel approach using Aspect-Based Sentiment Analysis (ABSA) to decode obfuscated sentiment in Thai financial annual reports. We develop specific guidelines for annotating obfuscated sentiment in these texts and annotate more than one hundred financial reports. We then benchmark various text classification models on this annotated dataset, demonstrating strong performance in sentiment classification. Additionally, we conduct an event study to evaluate the real-world implications of our sentiment analysis on stock prices. Our results suggest that market reactions are selectively influenced by specific aspects within the reports. Our findings underscore the complexity of sentiment analysis in financial texts and highlight the importance of addressing obfuscated language to accurately assess market sentiment.
\end{abstract}

\section{Introduction}

Human behavior in the financial world is highly complex, influenced by a multitude of factors, including market sentiment, macroeconomic indicators, and geopolitical events. These factors directly affect the demand and supply of stocks, thereby shaping market dynamics. For instance, positive earnings reports or favorable economic policies can drive up demand, while political instability or adverse regulatory changes can lead to a decline. In addition to these factors, annual reports mandated by government compliance offer another valuable source of signals for understanding market behavior. These reports contain rich textual information that reflects the underlying complexities of financial markets as demonstrated by the work of \cite{Chuthanondha2019}. Nevertheless, sentiment analysis of financial reports in the Thai language remains largely unexplored. However, despite the potential insights offered by such unstructured textual data, previous studies have predominantly focused on numerical data, overlooking the wealth of information embedded in text. We argue that this underexplored area presents a significant opportunity to enhance our understanding of financial markets through the systematic analysis of unstructured data.

Aspect-based sentiment analysis (ABSA) is a powerful natural language processing (NLP) technique that allows for a detailed understanding of sentiment by identifying specific aspects within a text and determining the sentiment associated with each. While ABSA has been widely studied and successfully applied to review data, where consumers openly express opinions on various aspects of products or services, its application to annual reports is notably more challenging. This difficulty stems from the fact that companies often frame their narratives in a consistently positive or neutral light, even when the underlying reality may be less favorable. For instance, a company might describe a significant reduction in its workforce as a "strategic realignment," thereby masking the potential negative sentiment with a positive or neutral expression. This practice introduces an additional layer of complexity to the analysis. The first layer involves addressing the linguistic variations used to express sentiment, while the second layer requires uncovering the company's intent to obscure negative sentiment through carefully chosen language. Together, these layers complicate the accurate application of ABSA to the nuanced and often guarded language found in annual reports.

In this project, we seek to analyze the aspect-based sentiment of Thai financial annual reports, which are mandated by the government and issued by companies publicly traded on the Thai stock market. To address the challenges posed by the often-obfuscated language in these reports, we establish guidelines for aspect-based sentiment annotation. These guidelines are specifically designed to delineate various aspects indicative of economic signals in financial documents and uncover the true sentiment of each aspect, even when the language is crafted to conceal negative sentiments. We then apply these guidelines in the annotation of over one hundred financial reports and achieve a high inter-annotator agreement score that confirms the consistency and reliability of our annotation process. 

Building on this foundation, we benchmark an array of text classification models on our annotated dataset and achieve strong accuracy in sentiment classification, demonstrating the effectiveness of our approach. Then we extend our analysis to assess the external validity of aspect-based sentiment analysis for Thai financial annual reports. We conduct an event study focused on the impact of annual report releases on stock prices to determine the real-world implications of our sentiment analysis. Our findings indicate that only specific aspects discussed in the documents correlate with abnormal returns, highlighting that market reactions are selective and influenced by particular elements within the reports. Furthermore, we observe that negative sentiment does not necessarily lead to an increase in returns after controlling for other variables, suggesting a complex and context-dependent relationship between sentiment expressed in annual reports and subsequent market performance. This underscores the importance of considering both the content of the reports and the broader market environment when interpreting the results of sentiment analysis. Consequently, our research questions are:

RQ1: How well can BERT-based models predict sentiment and aspect in financial reports where sentiment is deliberately obfuscated?

RQ2: Do sentiment and aspect that derived from official financial reports in the Thai language affect Thai stock market?

\section{Related Works}

Sentiment analysis refers to the analysis of emotions and aspects conveyed through various texts. A significant amount of research has been conducted and found that public sentiment and aspect, as well as information in financial reports, significantly influence various market indicators, such as prices \cite{Antweiler2004} and risks \cite{Wang2013}. Initially, researchers have utilized sentiment lexicons, repositories of words indicating emotions, to score texts as expressing positive, negative, or neutral sentiments \cite{Loughran2011, Sohangir2018}. Previous studies have focused on employing a variety of machine learning models along  with large dataset collected from X (formerly known as Twitter) to build a sentiment lexicon for applications on text from other sources \cite{Li2017, Oliveira2016}. Additionally, \citet{Loughran2020} emphasized the limitations of previous studies due to its biases in researchers’ criteria on selecting lexicons for analysis. They instead encourage the use of both supervised and unsupervised machine learning in future studies.

In the context of Thai stock market, the study of text analysis is still limited. Most of the studies depend on the analysis of text from online media or online news platforms \cite{Chatchawan2020, Tantisantiwong2020} more than official financial documents presented to regulatory authorities. Additionally, there is a related study  
that conducted a textual analysis on English-language financial disclosure of listed companies in Thailand \cite{Chuthanondha2019}. The finding indicates that indices derived from sentiment and aspect data can be utilized to predict returns or fluctuation in stock prices in Thai stock market.

Previous research has extensively applied Deep Learning for Financial Sentiment Analysis. \citet{Day2016} utilized a Deep Learning model to analyze financial news in Chinese. They created a feature vector based on the count of words in sentiment lexicon, without utilizing word embeddings. It resulted in suboptimal performance. \citet{Jangid2018} employed Recurrent Neural Network (RNN) in the form of Bidirectional Long-Short Term Memory (Bi-LSTM) and Convolutional Neural Network (CNN) for aspect-based financial sentiment analysis, achieving an F1 score of 0.69 for aspect classification and a mean-squared error of 0.112 for sentiment classification. Additionally. \citet{Akhtar2017} employed a similar model for prediction. 

To conduct sentiment analysis of stock market, \citet{Sousa2019} fine-tuned Bidirectional Encoder Representations from Transformers (BERT) on general domain sentiment analysis corpus and benchmarked on news article dataset.  News articles are often written by journalists and analysts who might not intentionally try to obfuscate the sentiment to present the positive sides of the organizations. This limits the applicability of the analysis. Moreover, \citet{Araci2019} trained a BERT model specifically for financial text analysis, achieving an F1 score of 0.84, surpassing LSTM models. \citet{Peng2021} further demonstrated that fine-tuning a pre-trained BERT model on large financial text data improved its performance by an additional 0.01 in F1 score.

\section{Financial Obfuscated Sentiment Dataset}

We prepare and annotate the dataset drawn from Thai annual financial reports. The dataset and the annotation guidelines are designed to uncover sentiment expressed by an obfuscated language.

\begin{table*}[ht]
\small
\centering
\begin{tabular}{lp{0.8\textwidth}}
\toprule
 Aspect & Description \\
\midrule
 Brand & Branding includes product logo, organizational image, and product image. \newline Marketing includes advertising, PR, marketing rewards, sales promotion, and sponsorship \\
\hline
 Product/Service & New product/service launches, product/service market testing  \newline Changes: Upgrades/downgrades, recalls, approvals, collaboration with other companies, licensing, alliances, partnerships, MOUs, Joint Ventures \\
\hline
 Environment & Environmental operations, environmental policies, environmental science, global warming, climate change, waste generation, pollution release, CSR activities related to the environment \\
\hline
 Social \& People & Hiring, changes in company staffing (hiring or firing), employee compensation, management changes (e.g., CEO, senior executives), work stoppages (strikes), CSR activities related to society, employees, labor, customers, communities, or other stakeholders \\
\hline
 Governance & Changes in the Board of Directors, company governance policies and subsidiary companies, transparency in operations, ethics, management oversight \\
\hline
 Economics & Economic Discourse that may affect a company, including country and global economic conditions, economic policies, international trade policies such as Free trade agreements (FTA), economic indices such as GDP, interest rates, inflation rates, unemployment rates, national income, exchange rates, economic trends in industries, countries, and globally \\
\hline
 Political & National or international political changes such as elections, coups, political movements, political unrest, wars, tax policies \\
\hline
 Legal & Legal disputes or decisions related to the law, investigations, allegations, lawsuits, litigation, prosecutions, verdicts, laws, and other legal issues \\
\hline
 Dividend & Payments made to company shareholders. Dividend payments may be in the form of cash, shares, or other assets. Changes related to dividend payments such as forecasts, reporting, announcements \\
\hline
 Investment & Capital expenditures in the company, subsidiaries, or joint ventures, investments in production infrastructure (e.g., factories), investments in products or services  \newline  Investments in research and development (R\&D)  \newline Events related to factories, office buildings, branches, warehouses, or other real estate, excluding M\&A \\
\hline
 M\&A & Merger and Acquisition (M\&A) activities of the company  \newline - Merger: When two or more companies merge to form a new company  
 \newline - Acquisition: When one company acquires some or all of the business of another company, divided into share acquisition and Asset \\
\hline
 Profit/Loss & Company performance, including revenue, sales, costs of goods sold, expenses, financial figures, or financial ratios.  Changes in stock prices in the securities market \\
\hline
 Rating & Company's creditworthiness rankings - ratings, rankings of organizational credibility, or rankings of the creditworthiness of each bond, reflecting the ability to repay debt securities. Credit rating by agencies in Thailand include TRIS Corporation (TRIS) and Fitch Ratings (Fitch), analyst recommendations (e.g., buy/sell/hold recommendations), including changes in recommendations \\
\hline
 Financing & Syndicated loan arrangements, issuance of bonds, capital raising in the securities market, stock repurchases, inter-company lending, IPO, private placement of shares, tender offer, increase/decrease in capital from VC or angel investors \\
\hline
 Technology & Technological changes, information technology, use of automation, use of AI , tech access, licensing, patents, and intellectual property rights \\
\hline
 Others &  Other changes beyond those mentioned above, such as disasters, pandemics \\
\bottomrule
\end{tabular}
\caption{Criteria on aspect annotation}
\label{tab-aspect-list}
\end{table*}

\subsection{Annotation Guidelines}
We annotate aspect and sentiment for each paragraph from selected sections in the document. We specify 16 aspects to cover most topics discussed in the financial reports that might be useful for predicting certain financial behavior in the capital market (Table \ref{tab-aspect-list}). The sentiment labels are negative, positive, and neutral. Each paragraph might be annotated with multiple aspect-sentiment pairs. 

\begin{figure*}[ht]
\centering
\includegraphics[width=0.9\textwidth]{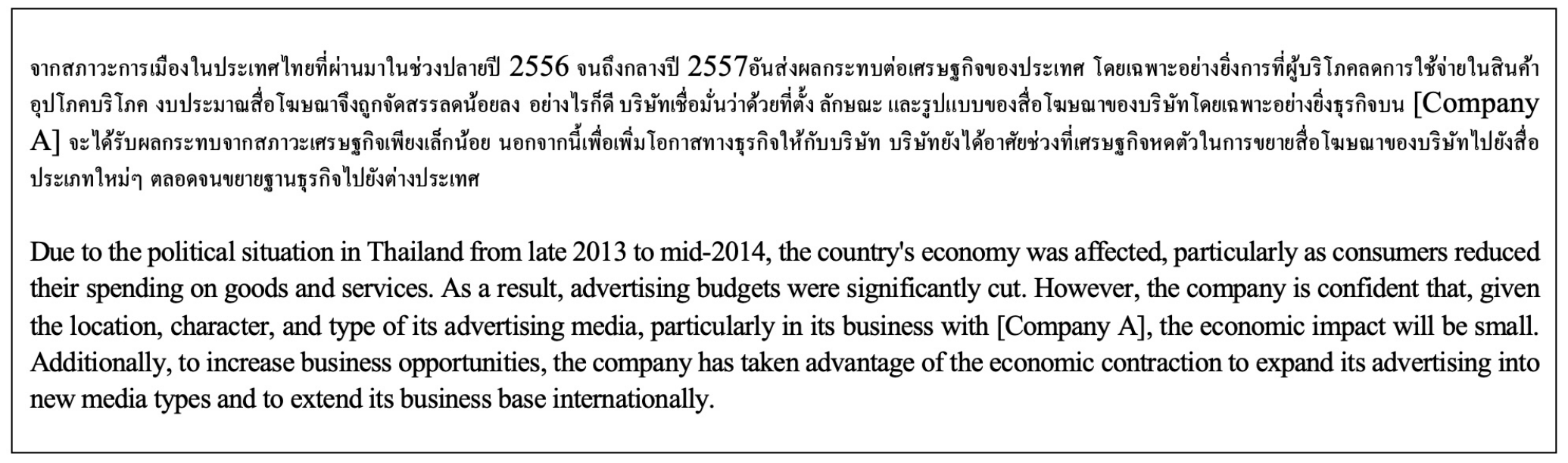}
\caption{Example of strategically positive framing obfuscating negative sentiment from financial stress (drawn from the risk section in the report}
\label{example1}
\end{figure*}

\begin{figure*}[ht]
\centering
\includegraphics[width=0.9\textwidth]{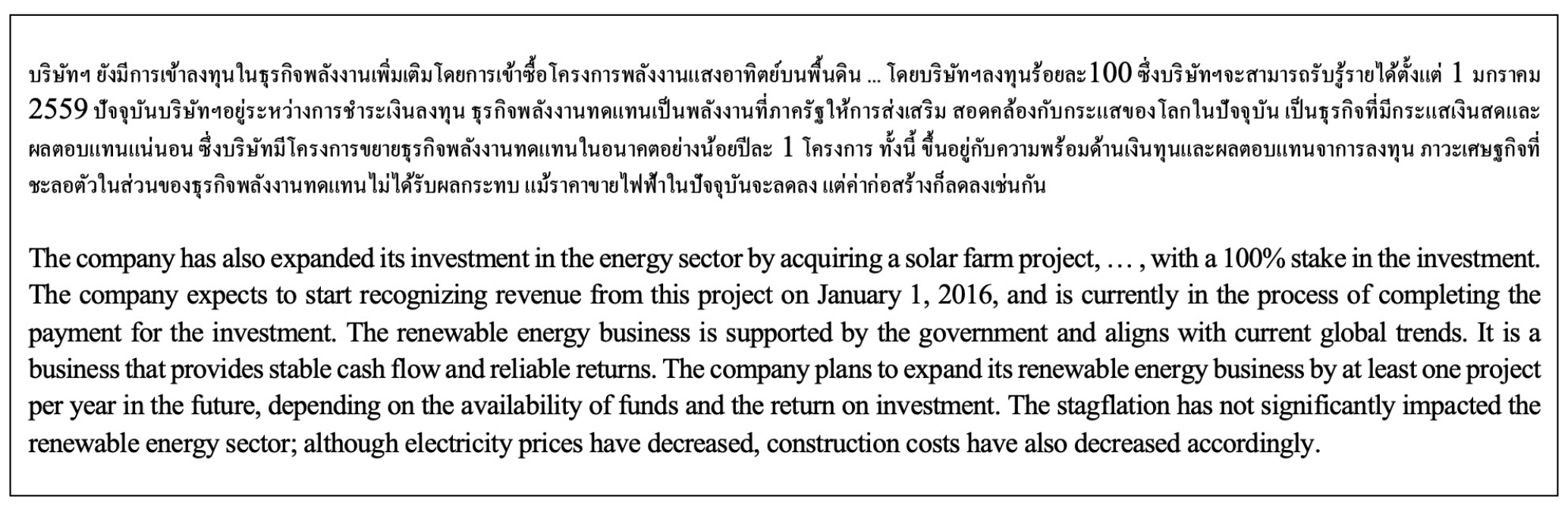}
\caption{Example of positive narrative obfuscating negative sentiment caused by uncertainty in investment decisions (drawn from the MD\&A section in the report}
\label{example2}
\end{figure*}

While aspect annotation is fairly straightforward, sentiment expressed in official financial reports show unique characteristics that cannot be adequately captured using typical sentiment annotation guidelines. For this reason, we call this annotation 
\emph{obfuscated sentiment}. General sentiment analysis often fails to reflect investor emotions accurately because many seemingly positive statements in reports are crafted to mitigate underlying negative implications for readers.  This phenomenon is illustrated in Figures \ref{example1} and \ref{example2}. 

Figure \ref{example1} outlines how the company faced financial difficulties due to political instability and responded by expanding into new advertising media. Although this expansion is framed as an optimistic strategy, it may also suggest that the company is struggling and reacting to economic pressures. Investors might perceive this move as a sign of financial instability, raising concerns about the company’s overall performance and prospects.

Taken from the paragraph located in the different section of another company, Figure \ref{example2} discusses the company's investment in renewable energy, presenting it as a positive strategic move supported by stable cash flow and government policy. However, it also acknowledges economic challenges like stagflation, which could impact the investment. The conditional future expansion based on funds and returns suggests uncertainty. This focus on positives might obscure underlying risks and raise concerns about the company’s ability to manage these challenges effectively.
The examples above illustrate that language used in official financial report often use optimistic language to frame the company's strategies positively while subtly redirecting attention from underlying risks and challenges. Therefore, the creation of a financial sentiment dataset requires careful attention to develop new sentiment guidelines.

\begin{table}[ht]
\centering
\scriptsize
\begin{tabular}{llll}
\toprule
 Aspect & Train & Dev & Test \\
\midrule
 Brand & 113 (1.38\%) & 113 (1.38\%) & 113 (1.38\%) \\
 Product/Service & 47 (2.68\%) & 47 (2.68\%) & 47 (2.68\%) \\
 Environment & 29 (1.65\%) & 29 (1.65\%) & 29 (1.65\%) \\
 Social \& People & 894 (10.9\%) & 894 (10.9\%) & 894 (10.9\%) \\
 Governance & 243 (13.84\%) & 243 (13.84\%) & 243 (13.84\%) \\
 Economics & 159 (9.06\%) & 159 (9.06\%) & 159 (9.06\%) \\
 Political & 491 (5.99\%) & 491 (5.99\%) & 491 (5.99\%) \\
 Legal & 150 (8.54\%) & 150 (8.54\%) & 150 (8.54\%) \\
 Dividend & 64 (0.78\%) & 64 (0.78\%) & 64 (0.78\%) \\
 Investment & 7 (0.45\%) & 7 (0.45\%) & 7 (0.45\%) \\
 M\&A & 13 (0.74\%) & 13 (0.74\%) & 13 (0.74\%) \\
 Profit/Loss & 283 (3.46\%) & 283 (3.46\%) & 283 (3.46\%) \\
 Rating & 81 (4.61\%) & 81 (4.61\%) & 81 (4.61\%) \\
 Financing & 81 (4.62\%) & 81 (4.62\%) & 81 (4.62\%) \\
 Technology & 37 (0.45\%) & 37 (0.45\%) & 37 (0.45\%) \\
 Others & 19 (1.08\%) & 19 (1.08\%) & 19 (1.08\%) \\
\bottomrule
\end{tabular}
\caption{The distribution of aspect labels}
\label{tab-aspect-distribution}
\end{table}

\begin{table}[htbp]
\scriptsize
\centering
\begin{tabular}{llll}
\toprule
 Sentiment & train  & dev  & test  \\
\midrule
 Negative & 1,316 (16.07\%) & 226 (15.15\%) & 323 (18.40\%) \\
 Neutral & 3,751 (45.79\%) & 686 (39.07\%) & 821 (46.78\%) \\
 Positive & 3,124 (38.14\%) & 804 (45.79\%) & 611 (34.81\%) \\
\bottomrule
\end{tabular}
\caption{The distribution of sentiment labels}
\label{tab-sentiment-distribution}
\end{table}

\subsection{Data Annotation and Statistics}
All annotations were conducted by four annotators with backgrounds in economics, ensuring that the interpretations of financial language were grounded in domain expertise. Before beginning large-scale annotation, the annotators underwent a training phase in which they practiced on sample paragraphs and iteratively refined their understanding of the guidelines. This training ensured that all annotators applied the aspect and sentiment criteria consistently. The average inter-annotator Cohen’s kappa score reached 0.73 for aspect annotation and 0.77 for sentiment annotation, both falling within the \textit{substantial agreement} range \cite{landis1977measurement}. These results demonstrate that, despite the challenges posed by obfuscated and strategically framed corporate language, the annotation guidelines successfully enabled reliable and consistent labeling across all annotators.

We divide the dataset into three sets: a training set comprising 8,191 paragraphs (70\%), a development set consisting of 1,756 paragraphs (15\%), and a test set containing 1,755 paragraphs (15\%). The training set comprises paragraphs extracted from Form 56-1 for the years 2015 to 2018 for selected companies. The validation set includes paragraphs from Form 56-1 for the year 2019. The test set encompasses paragraphs from Form 56-1 for the years 2018 to 2019. 

The distribution of aspects and sentiments within the annotated Form 56-1 dataset reveals a balanced representation across different categories. Aspects such as "Social \& People" and "Governance" are more frequently annotated, highlighting their importance in the reports (Table \ref{tab-aspect-distribution}). In contrast, less common aspects like "Investment" and "M\&A" appear less frequently, but the overall distribution remains balanced. The sentiment distribution reveals neutral sentiment labels to be the majority, but both positive and negative sentiments are also well-represented (Table \ref{tab-sentiment-distribution}). This balanced distribution suggests that the reports contain a mix of content, with a slight emphasis on neutrality, typical of corporate disclosures. The even distribution across sentiments and aspects indicates that the dataset is sufficiently conducive to developing models that can generalize well across different labels.

The annotated dataset, including aspects, sentiment labels, and annotation guidelines, is publicly available at \url{https://github.com/nlp-chula/finnlp-sentiment} to support future research.

\section{Experiment Setting}
In the analysis of sentiment and aspect within the financial report, we consider that models should address the problem in the form of a multi-class classification. This means that one paragraph can predict one label from more than two types of labels. Although one paragraph may involve multi-label classification, where there can be multiple sentiment or aspect labels, the performance of baseline models was relatively close to those of the multi-class classification. We, therefore, confines the scope of the study to the multi-class classification problem only. In addition, we divided models into two experiments. The first experiment is models for sentiment classification. Another is models for aspect classification.

In both experiments, we fine-tune a language model pretrained on Thai-language general data (Pretrained Language Model: PLM) called WangchanBERTa \cite{wangchanberta} with our tailored aspect-sentiment dataset to address our research question regarding the contextualized meaning in writing style of financial report. For the baseline models, Maximum Entropy (MaxEnt), and Convolutional Neural Network (CNN) models were employed. 

As preprocessing steps, we use pythainlp library default word tokenizer to tokenize the text. We remove punctuation, English characters, numerals, and tokens with a character length of less than three. MaxEnt models use bag-of-word count features.

We test a CNN model with the expectation of capturing the phrasal structure of text using word embeddings of consecutive words. In this setup, we tokenize the data using the Attacut engine \cite{chormai-etal-2020-syllable}  and constrain the maximum length of tokens to 100. The word embeddings we utilize are from the Universal Sentence Encoder. For aspect classification, we configure the model with 500 filters, a kernel size of 4, and hidden dimensions of 100. Additionally, a dropout rate of 0.1 is applied. The model operates with a batch size of 32 and undergoes training for 10 epochs. For sentiment classification, the configuration remains identical expect for a change in hidden dimensions to 200.

We select the ‘wangchanberta-base-att-spm-uncased’ version of the pretrained language model from huggingface, which utilizes a BERT-based architecture. We then fine-tuned this model on our newly annotated dataset. For aspect prediction, we configured the model with a learning rate of 0.00003, a batch size of 16, trained it for 5 epochs, and applied a weight decay of 0.01. For sentiment prediction model, the learning rate was adjusted to 0.00005, while the other parameters remained consistent with those of the aspect classifier.

\begin{table}[ht]
\centering
\scriptsize
\begin{tabular}{lccc}
\toprule
\textbf{Aspect} & \textbf{MaxEnt} & \textbf{CNN} & \textbf{WangchanBERTa} \\
\midrule
Brand & 0.60 & 0.33 & \textbf{0.67} \\
Dividend & \textbf{0.92} & 0.73 & 0.90 \\
Economics & \textbf{0.79} & 0.78 & \textbf{0.79} \\
Environment & 0.80 & 0.76 & \textbf{0.86} \\
Financing & 0.57 & \textbf{0.64} & 0.57 \\
Governance & 0.78 & 0.77 & \textbf{0.79} \\
Investment & 0.59 & 0.59 & \textbf{0.62} \\
Legal & 0.58 & 0.55 & \textbf{0.67} \\
M\&A & 0.25 & 0.00 & \textbf{0.38} \\
Others & 0.61 & 0.60 & \textbf{0.71} \\
Political & 0.62 & \textbf{0.72} & 0.60 \\
Product/Service & 0.63 & 0.65 & \textbf{0.84} \\
Profit/Loss & 0.82 & 0.85 & \textbf{0.87} \\
Rating & \textbf{0.40} & 0.00 & 0.00 \\
Social \& People & 0.80 & 0.82 & \textbf{0.86} \\
Technology & 0.38 & 0.38 & \textbf{0.74} \\
\midrule
Accuracy (overall) & 0.74 & 0.74 & \textbf{0.79} \\
Micro avg (F1) & 0.74 & 0.74 & \textbf{0.79} \\
Macro avg (F1) & 0.63 & 0.57 & \textbf{0.66} \\
\bottomrule
\end{tabular}
\caption{WangchanBERTa substantially improves aspect classification—especially in subtle, low-frequency categories}
\label{tab-aspect-f1}
\end{table}

\begin{table}[ht]
\centering
\scriptsize
\begin{tabular}{lccc}
\toprule
\textbf{Sentiment} & \textbf{MaxEnt} & \textbf{CNN} & \textbf{WangchanBERTa} \\
\midrule
Negative & 0.65 & 0.67 & \textbf{0.74} \\
Neutral & 0.75 & 0.78 & \textbf{0.78} \\
Positive & 0.70 & 0.68 & \textbf{0.77} \\
\midrule
Accuracy (overall) & 0.72 & 0.73 & \textbf{0.77} \\
Micro avg (F1) & 0.72 & 0.73 & \textbf{0.77} \\
Macro avg (F1) & 0.70 & 0.71 & \textbf{0.76} \\
Weighted avg (F1) & \textbf{0.72} & 0.72 & 0.77 \\
\bottomrule
\end{tabular}
\caption{Contextual modeling yields significant performance gains in sentiment classification under obfuscation.}
\label{tab-sentiment-f1}
\end{table}

\section{Results and Discussion}
\textbf{RQ1: How well can BERT-based models predict sentiment and aspect in financial reports where sentiment is deliberately obfuscated?} The performance of WangchanBERTa suggests that sentiment classification benefits from considering sentence context, along with keywords (Table \ref{tab-sentiment-f1}). The WangchanBERTa model achieves a high 79\% accuracy, outperforming models relying solely on keywords like Maximum Entropy (74\%). This indicates the importance of paragraph-level context for efficient sentiment analysis. Adding context beyond keywords improves accuracy by 5\%. Despite having to operate 16-way classification, both BERT-based models show good effectiveness in real-world data aggregation. The CNN model, which tests the efficiency of using very narrow contextual information, does not perform better than models using keywords for aspect classification. This suggests that expanding the contextual consideration to at least the sentence level is necessary to achieve higher accuracy in aspect classification.

WangchanBERTa achieves a good 77\% accuracy, which is considered high compared to similar Thai language datasets (Table \ref{tab-sentiment-f1}). This suggests that the text from Form 56-1 may have common patterns across companies. The most efficient model is WangchanBERTa, followed by CNN and Maximum Entropy, respectively. CNN, which utilizes local context, performs better than Maximum Entropy in extracting signals for sentiment classification. This illustrates that using keywords alone is insufficient for sentiment analysis, emphasizing the need for broader context. WangchanBERTa can consider a broader linguistic context than CNN due to its use of self-attention, which can analyze dependencies between distant constituents of a sentence. Therefore, WangchanBERTa is the most suitable model for this task.

\section{Event Study}
We want to study the effects of financial reports on their respective companies’ stock prices. The study of impacts of a particular event on firm’s stock is called 'event study' \cite{mackinlay1997event}. This study relies on the assumption that capital market reflects the information about the firm in company’s stock price. To quantify an event’s economic impact, abnormal returns are calculated by deducting the normal returns, which would have been realized if the analyzed event has not occurred, from the actual return. We had tried estimating the normal returns from three expected return models: (1) constant mean model using the average return of the stock over the past year, (2) market model utilizing linear regression modeling of the SET market index to calculate returns over the past year, and (3) Fama-French model consisting of five variables: Mkt-RF (Market), SMB (Size), HML (Value), RMW (Operating Profitability), and CMA (Investment). Finally, the abnormal return deducted from the output of Fama-French model was selected to be used in regression analysis in the next section, since there was no significant difference observed when comparing the results of each model.

In this context, the analyzed event is designed to be one day after the submitted date of Form 56-1 via SETLink, a portal for document submission to The Stock Exchange of Thailand, by companies. According to SET regulations, financial reports are published the day after their submission. Our study focuses on 189 non-financial companies listed in the SET Index or MAI Index, spanning returns from 2015 to 2022. The methodology involves three sets of event windows: ±5 days, ±3 days, and ±1 day, with an estimation size of 250 days. 

\subsection{Regression Analysis}
To investigate the impact of sentiment extracted from Thai financial reports on the Thai stock markets, a linear regression model was utilized. The model aimed to confirm the correlation between sentiment expressed in various aspects, treated as predictors, and cumulative average abnormal return (CAAR), the dependent variable. The dataset encompasses 700 observations pertaining to 189 firms listed on the Stock Exchange of Thailand (SET) for the period 2015 to 2022. Regarding the sentiment, we extract sentiments and aspects from the financial reports of those companies by utilizing the optimal language models from previous sections. The automatic annotation of aspect-based sentiment was done by the fine-tuned WangchanBERTa model. 

We structure this experiment as a cross-sectional analysis based on the industries in which companies are registered to mitigate overlapping event windows across companies, which could possibly lead to clustering problem in statistical analysis. The event windows are calculated with the referenced event date as the release date of the document plus one day (t+1). The regression models incorporate five firm-level control variables to account for financial and operational characteristics that may influence the dependent variable. The variables are defined as follows:

\begin{itemize}
    \item \textbf{Firm size}: measured as the natural logarithm of the firm's market capitalization.
    \item \textbf{Tobin's Q}: calculated as the natural logarithm of the ratio of market capitalization to total assets.
    \item \textbf{Return on assets (ROA)}: the ratio of net income to total assets.
    \item \textbf{Leverage}: the ratio of total liabilities to total assets.
    \item \textbf{Volatility}: the standard deviation of firm returns, estimated from the 12 months preceding the event date.
\end{itemize} 

For each observation, sentiment is represented using either one-hot embedding or sentiment score. The main regression predictors are derived from the textual analysis of financial documents. Each variable is defined below:

\begin{itemize}
    \item \textbf{Aspect}: a categorical variable representing the type of information in each sentence, with 16 possible types as described earlier.
    \item \textbf{Sentiment}: a categorical variable indicating emotional polarity with three classes: negative, neutral, and positive.
    \item \textbf{Source}: the section of the financial document from which a sentence is extracted; three sources are used: Management Discussion \& Analysis (MDA), Sustainability, and Risk.
    \item \textbf{Industry}: a set of dummy variables representing the firm's industry classification, with eight broad industry categories defined by the Stock Exchange of Thailand (SET).
    \item \textbf{Score adapted from Tan et al. (2015)}: a normalized sentiment score computed using the formula presented in the methodology.
\end{itemize}

{\small
\noindent\textbf{Model 1:}
\begin{multline}
\label{eq:model1}
CAAR = \beta_0 + \beta_1(\text{Sentiment}) + \beta_2(\text{Controls})\\
+ \beta_3(\text{Industry})
\end{multline}

\noindent\textbf{Model 2:}
\begin{multline}
\label{eq:model2}
CAAR = \beta_0 + \beta_1(\text{Sentiment}) + \beta_2(\text{Controls})\\
+ \beta_3(\text{Industry}) + \beta_4(\text{Score})
\end{multline}

\noindent\textbf{Model 3:}
\begin{multline}
\label{eq:model3}
CAAR = \beta_0 + \beta_1(\text{Source.Sentiment}) + \beta_2(\text{Controls})\\
+ \beta_3(\text{Industry})
\end{multline}

\noindent\textbf{Model 4:}
\begin{multline}
\label{eq:model4}
CAAR = \beta_0 + \beta_1(\text{Source.Sentiment}) + \beta_2(\text{Controls})\\
+ \beta_3(\text{Industry}) + \beta_4(\text{Score})
\end{multline}

\noindent\textbf{Model 5:}
\begin{multline}
\label{eq:model5}
CAAR = \beta_0 + \beta_1(\text{Aspect.Sentiment}) + \beta_2(\text{Controls})\\
+ \beta_3(\text{Industry})
\end{multline}
}


In addition to ordinary least square (OLS) linear regression, we also use ridge regression model (also known as L2 regularization) to prevent multicollinearity and overfitting due to the inclusion of numerous correlated predictors. We apply bootstrap row resampling to obtain standard errors of regression coefficients from 10,000 resamples \cite{efron2000bootstrap} since classical statistical tests on coefficients no longer apply to ridge regression.

\begin{table}[ht]
\centering
\small
\begin{tabular}{lcc}
\toprule
\textbf{Predictor Variable} & \textbf{Model 1} & \textbf{Model 2} \\
\midrule
Negative   & 0.000200        & 0.000300$^{*}$ \\
Neutral    & -0.000014       & -0.000016$^{*}$ \\
Positive   & 0.000029        & 0.000042 \\
Score-1 = (p-n+1)/(p+n) & -- & -0.050800 \\
Score-2 = (p-n+2)/(p+n) & -- & 0.050800 \\
Industry   & Y & Y \\
Controls   & Y & Y \\
Intercept  & -0.010600 & -0.012700 \\
\midrule
R-squared  & 0.17 & 0.17 \\
\bottomrule
\end{tabular}
\caption{Regression results at window size [-3,3] for Model 1 and Model 2. $^{*}$ and $^{**}$ represent 5\% and 1\% statistical significance, respectively. Y = included in the model. Numbers in parentheses are standard errors.}
\label{tab-model-1-2}
\end{table}

\begin{table}[ht]
\centering
\small
\begin{tabular}{lcc}
\toprule
\textbf{Predictor Variable} & \textbf{Model 3} & \textbf{Model 4 } \\
\midrule
\textit{Negative} \\
MDA & 0.000400 & 0.000500$^{*}$ \\
Risk & 0.000200 & 0.000200 \\
Sustainability & -0.000079 & -0.000055 \\
\midrule
\textit{Neutral} \\
MDA & -0.000019 & -0.000021 \\
Risk & -0.000010 & -0.000012 \\
Sustainability & 0.000002 & 0.000000 \\
\midrule
\textit{Positive} \\
MDA & -0.000001 & 0.000003 \\
Risk & 0.000015 & 0.000028 \\
Sustainability & 0.000036 & 0.000055 \\
\midrule
Score-1 = (p-n+1)/(p+n) & -- & -0.051700 \\
Score-2 = (p-n+2)/(p+n) & -- & 0.051700 \\
Industry & Y & Y \\
Controls & Y & Y \\
Intercept & -0.013600 & -0.015800 \\
\midrule
R-squared & 0.17 & 0.18 \\
\bottomrule
\end{tabular}
\caption{Regression results at window size [-3,3] for Model 3 and Model 4. $^{*}$ and $^{**}$ represent 5\% and 1\% statistical significance, respectively. Y = included in the model. Numbers in parentheses are standard errors.}
\label{tab-model-3-4}
\end{table}

\subsection{Results and Discussion}
\textbf{RQ2: Do sentiment and aspect that derived from official financial reports in Thai language affect Thai stock market?} Our regression analysis reveals a nuanced relationship between sentiment expressed in Form 56-1 reports and subsequent market reactions. When using only aggregated sentiment categories (Model 1), none of the coefficients reach statistical significance, suggesting that simple polarity labels (positive/neutral/negative) alone are insufficient to explain abnormal returns (Table \ref{tab-model-1-2}). However, once we incorporate the normalized sentiment scores (Score-1/Score-2) in Model 2, both negative and neutral sentiment become statistically significant ($p < 0.05$). Interestingly, negative sentiment is associated with higher returns, while neutral sentiment corresponds to lower returns. This suggests that the market interprets weakly framed, cautious, or "neutral-sounding" statements more negatively than explicit negative disclosures potentially because neutral framing often masks unfavorable realities in obfuscated financial language.

To further examine where sentiment matters, Models 3 and 4 separate sentiment by document section (Table \ref{tab-model-3-4}). Only negative sentiment originating from the MD\&A section shows significance across specifications. This highlights the informational importance of management’s own narrative; investors appear to react more strongly to adverse sentiment when it is voiced directly by the company’s executives rather than in other sections like Risk or Sustainability.


\begin{table*}[ht]
\centering
\tiny

\begin{tabular}{llrrrrrrrrr}
\toprule
\multicolumn{2}{c}{Predictor variables} &
\multicolumn{3}{c}{CAAR [-5,5]} &
\multicolumn{3}{c}{CAAR [-3,3]} &
\multicolumn{3}{c}{CAAR [-1,1]} \\
\cmidrule(lr){1-2}\cmidrule(lr){3-5}\cmidrule(lr){6-8}\cmidrule(lr){9-11}
& &
Coef. & Std.Error & Sig. &
Coef. & Std.Error & Sig. &
Coef. & Std.Error & Sig. \\
\midrule
Brand           & Positive & 0.000448  & 0.00106  &     & 0.000264  & 0.00082  &     & 0.000255  & 0.00045  &     \\
Brand           & Negative & 0.018574  & 0.008255 & *   & 0.006900  & 0.006865 &     & 0.001127  & 0.004995 &     \\
Product/Service & Positive & 0.000149  & 0.000201 &     & 0.000071  & 0.000155 &     & 0.000108  & 0.000124 &     \\
Product/Service & Neutral  & -0.000028 & 0.000068 &     & -0.000063 & 0.000051 &     & -0.000066 & 0.000038 &     \\
Product/Service & Negative & -0.001853 & 0.001883 &     & -0.000327 & 0.002439 &     & 0.000303  & 0.001274 &     \\
Environment     & Positive & -0.00011  & 0.000335 &     & 0.000107  & 0.000278 &     & -0.000018 & 0.000236 &     \\
Environment     & Neutral  & -0.000031 & 0.000146 &     & 0.000140  & 0.000114 &     & 0.000112  & 0.000082 &     \\
Environment     & Negative & 0.001561  & 0.002832 &     & -0.000531 & 0.002439 &     & -0.000442 & 0.001946 &     \\
Social\&People  & Positive & -0.000273 & 0.000297 &     & -0.000424 & 0.000235 &     & -0.000212 & 0.000156 &     \\
Social\&People  & Neutral  & 0.000165  & 0.000151 &     & 0.000251  & 0.000132 &     & 0.000174  & 0.000095 &     \\
Social\&People  & Negative & 0.000299  & 0.002269 &     & 0.001951  & 0.001804 &     & 0.000364  & 0.001219 &     \\
Governance      & Positive & -0.001188 & 0.000731 &     & -0.000430 & 0.000565 &     & -0.000108 & 0.000354 &     \\
Governance      & Neutral  & 0.000012  & 0.000064 &     & -0.000070 & 0.000057 &     & -0.000077 & 0.000049 &     \\
Governance      & Negative & 0.001719  & 0.002328 &     & 0.002579  & 0.001753 &     & 0.000788  & 0.001393 &     \\
Economics       & Positive & -0.000025 & 0.001447 &     & 0.000003  & 0.00112  &     & 0.001609  & 0.001287 &     \\
Economics       & Neutral  & -0.001636 & 0.001283 &     & -0.001761 & 0.000057 & **  & -0.001058 & 0.000789 &     \\
Economics       & Negative & 0.000311  & 0.001583 &     & 0.000983  & 0.000912 &     & 0.000308  & 0.000749 &     \\
Political       & Positive & 0.000854  & 0.004528 &     & -0.001403 & 0.003736 &     & 0.000479  & 0.002649 &     \\
Political       & Neutral  & 0.001295  & 0.003491 &     & 0.002104  & 0.002952 &     & -0.000152 & 0.002285 &     \\
Political       & Negative & -0.007455 & 0.006603 &     & -0.008209 & 0.005304 &     & -0.004415 & 0.00371  &     \\
Dividend        & Positive & -0.000325 & 0.00158  &     & 0.000496  & 0.001395 &     & -0.000812 & 0.000801 &     \\
Dividend        & Neutral  & 0.002502  & 0.002419 &     & 0.003384  & 0.002    &     & 0.00176   & 0.001172 &     \\
Dividend        & Negative & -0.000513 & 0.006244 &     & 0.000378  & 0.004447 &     & -0.001239 & 0.003508 &     \\
Legal           & Positive & 0.002023  & 0.002865 &     & 0.001189  & 0.001948 &     & -0.000081 & 0.001567 &     \\
Legal           & Neutral  & -0.000485 & 0.001121 &     & 0.000036  & 0.000912 &     & 0.000428  & 0.000824 &     \\
Legal           & Negative & -0.000114 & 0.002455 &     & 0.000285  & 0.002416 &     & 0.001303  & 0.000824 &     \\
Investment      & Positive & 0.000108  & 0.001641 &     & 0.000245  & 0.001196 &     & 0.000359  & 0.000964 &     \\
Investment      & Neutral  & -0.000057 & 0.000633 &     & 0.000217  & 0.000436 &     & 0.000027  & 0.000325 &     \\
Investment      & Negative & -0.006359 & 0.006921 &     & -0.008940 & 0.00574  &     & -0.004742 & 0.003523 &     \\
M\&A            & Positive & -0.002165 & 0.004929 &     & 0.001501  & 0.003789 &     & 0.001369  & 0.002401 &     \\
M\&A            & Neutral  & -0.003364 & 0.005209 &     & -0.002646 & 0.003606 &     & -0.005671 & 0.003948 &     \\
M\&A            & Negative & -0.027279 & 0.026297 &     & -0.017885 & 0.017196 &     & -0.001376 & 0.009132 &     \\
Profit/Loss     & Positive & -0.000601 & 0.000361 &     & -0.000926 & 0.000308 & **  & -0.000727 & 0.00025  & **  \\
Profit/Loss     & Neutral  & -0.000029 & 0.000087 &     & -0.000077 & 0.000072 &     & -0.000063 & 0.000053 &     \\
Profit/Loss     & Negative & 0.00136   & 0.000527 & **  & 0.001764  & 0.000462 & **  & 0.001474  & 0.000442 & **  \\
Others          & Positive & 0.00392   & 0.001708 & *   & 0.002323  & 0.001369 &     & 0.000516  & 0.001036 &     \\
Others          & Neutral  & -0.000208 & 0.000186 &     & -0.000134 & 0.000153 &     & 0.000053  & 0.000105 &     \\
Others          & Negative & 0.002111  & 0.001561 &     & 0.000158  & 0.00138  &     & -0.001131 & 0.001129 &     \\
Financing       & Positive & -0.001466 & 0.002631 &     & -0.001955 & 0.001845 &     & -0.001166 & 0.001487 &     \\
Financing       & Neutral  & 0.000166  & 0.000412 &     & 0.000113  & 0.000336 &     & 0.000264  & 0.000329 &     \\
Financing       & Negative & -0.002085 & 0.002124 &     & -0.000116 & 0.001806 &     & 0.00031   & 0.001407 &     \\
Technology      & Positive & 0.000194  & 0.00225  &     & 0.002940  & 0.001982 &     & -0.000246 & 0.001319 &     \\
Technology      & Neutral  & -0.001308 & 0.001744 &     & -0.001223 & 0.001548 &     & 0.000084  & 0.000966 &     \\
Technology      & Negative & 0.007731  & 0.013522 &     & 0.011261  & 0.011137 &     & 0.006095  & 0.007838 &     \\
\midrule
R-squared       &           & 0.15     &          &     &  0.24     &           &    & 0.25     & \\
\bottomrule
\end{tabular}

\caption{The results of regression analysis over many different window sizes (Model 5) suggest that a small subset of aspect–sentiment combinations meaningfully predicts market reactions, with Profit/Loss (Negative) consistently driving CAAR.}
\label{tab:caar_ridge}

\end{table*}

\begin{table}
\centering
\begin{tabular}{crr}
\toprule
  & \textbf{OLS} & \textbf{Ridge}  \\
  \midrule
\textbf{CAAR {[}-5,5{]}} & 0.15                       & 0.11                      \\
\textbf{CAAR {[}-3,3{]}} & 0.241                      & 0.16                      \\
\textbf{CAAR {[}-1,1{]}} & 0.51                       & 0.25 \\
\bottomrule      
\end{tabular}
\caption{The goodness of fit increases in shorter windows, confirming that market reactions to textual sentiment are immediate but short-lived.}
\label{tab:carr_ols_ridge_r2}
\end{table}

Does aspect-based sentiment from financial report affect the stock prices upon release? The results suggest that only certain aspects have significant implications for the market (Table \ref{tab-impact}). We found that only negative sentiment from the Profit/Loss aspect is statistically significant throughout all window sizes. The coefficients of the model predictors corresponding to aspect-based sentiment are relatively low, indicating that the positive/negative sentiment in each aspect has a small impact on market returns. It is possible that the impact on the market from news and sentiments embedded in Form 56-1 in any sections has already been represented in the prices, as investors and the market have anticipated.

Our result further suggest that the market rapidly incorporates new financial performance signals at the report’s release. The R-squared values rise from relatively low levels in the ±5-day window to substantially higher values in the ±1-day window especially under OLS, where the model explains over half of the variation in CAAR (Table \ref{tab:carr_ols_ridge_r2}. This pattern suggests that investors incorporate textual signals from annual reports almost instantly once the documents are released (t+1), but these effects fade as the window expands and other market signals begin to dominate. In short, sentiment extracted from financial disclosures matters most in the very short run, reinforcing the view that Form 56-1 contributes to immediate information updating rather than sustained price movement. 


\begin{table}[ht]
\centering
\small
\begin{tabular}{ccc}
\toprule
\textbf{Rank} & \textbf{Positive impact} & \textbf{Negative impact} \\
\midrule
1 & Technology – Negative & M\&A – Negative \\
2 & Brand – Negative & Investment – Negative \\
3 & Dividend – Neutral & Politics – Negative \\
4 & Technology – Negative & M\&A – Neutral \\
5 & Governance – Negative & Financing – Positive \\
\bottomrule
\end{tabular}
\caption{Ranking of aspect-based sentiment impact on stock market during window size $[-3,+3]$}
\label{tab-impact}
\end{table}

Which aspect-based sentiment affects the stock prices the most? We rank the highest and the lowest coefficients that are statistically significant (Table \ref{tab-impact}). On the positive impact side, it is interesting to note that negative sentiments associated with "Technology," "Brand," and "Governance" are ranked as having the most significant positive effects on the market. This suggests that despite negative sentiment in these areas, the market may perceive them differently, possibly as opportunities for growth or recovery. Conversely, aspects like "M\&A" (Mergers and Acquisitions), "Investment," and "Politics" have negative impacts on the market, with a notable predominance of negative sentiment. Even when "M\&A" has neutral sentiment, it still has a negative impact on the market, which could indicate market wariness toward these activities. Similarly, "Financing" with positive sentiment also contributes negatively, suggesting a counterintuitive market response to positive financial news. This table highlights the complexity of sentiment analysis in market reactions, as sentiment polarity does not always align with market outcomes. 

\section{Conclusion}
Despite the challenges faced by existing research in financial sentiment analysis, such as outdated methods, limited exploration of advanced natural language processing techniques, and a focus on English text over Thai in studies of the Thai stock market, our study address these issues from data collection to model development. We create the first aspect-sentiment financial dataset tailored specifically to Thai financial reports. We adopt BERT-based models to predict aspects and sentiment in financial annual reports from companies in Thailand. Our model outperforms traditional keyword-based models like MaxEnt and even yields better results than narrow-context model like CNN, emphasizing the significance of broader context required to extract indirect sentiment expression. 

Moreover, our study highlights the significance of financial documents, particularly Form 56-1 or annual reports, in influencing the Thai stock market. This is because it introduces new information and reduces data asymmetry issues in the stock market. However, the impact on the sentiment and perception of information in the Thai stock market, as revealed by the regression analysis, suggests that a company's information affects investor sentiment and perceptions of the market's performance differently based on the aspect. Furthermore, the impact of sentiment on the market from certain aspect may not align with expectations according to financial theory.

Our dataset and resources have been released at \url{https://github.com/nlp-chula/finnlp-sentiment} to facilitate continued research in financial NLP for the Thai market.

\section*{Acknowledgments}
We would like to thank Capital Market Development Fund who support this research entitled Sentiment Analysis on Financial Documents under Grant CMDF-0062\_2566.

\bibliography{custom}

\appendix

\section{Appendix}
\label{sec:appendix}

\subsection{List of firms in the dataset}

\begin{itemize}\setlength\itemsep{0pt}
  \item AAV -- Asia Aviation Public Company Limited
  \item ACC -- Advanced Connection Corporation Public Company Limited
  \item ADVANC -- Advanced Info Service Public Company Limited
  \item AGE -- Asia Green Energy Public Company Limited
  \item AIT -- Advanced Information Technology Public Company Limited
  \item AJ -- A.J. Plast Public Company Limited
  \item AMC -- Asia Metal Public Company Limited
  \item AP -- AP (Thailand) Public Company Limited
  \item APCO -- Asian Phytoceuticals Public Company Limited
  \item APCS -- Asia Precision Public Company Limited
  \item AS -- Asphere Innovations Public Company Limited
  \item ASEFA -- Asefa Public Company Limited
  \item BA -- Bangkok Airways Public Company Limited
  \item BANPU -- Banpu Public Company Limited
  \item BEM -- Bangkok Expressway and Metro Public Company Limited
  \item BEYOND -- Bound and Beyond Public Company Limited
  \item BJC -- Berli Jucker Public Company Limited
  \item BJCHI -- BJCHI Corporation Public Company Limited
  \item BRR -- Buriram Sugar Public Company Limited
  \item BSBM -- Bangsaphan Barmill Public Company Limited
  \item BTS -- BTS Group Holdings Public Company Limited
  \item CBG -- Carabao Group Public Company Limited
  \item CCET -- Cal-Comp Electronics (Thailand) Public Company Limited
  \item CEN -- Capital Engineering Network Public Company Limited
  \item CENTEL -- Central Plaza Hotel Public Company Limited
  \item CGD -- Country Group Development Public Company Limited
  \item CITY -- City Steel Public Company Limited
  \item CKP -- CK Power Public Company Limited
  \item CM -- Chiangmai Frozen Foods Public Company Limited
  \item CPALL -- CP ALL Public Company Limited
  \item CPI -- Chumporn Palm Oil Industry Public Company Limited
  \item CPN -- Central Pattana Public Company Limited
  \item FORTH -- Forth Corporation Public Company Limited
  \item GLOBAL -- Siam Global House Public Company Limited
  \item GPSC -- Global Power Synergy Public Company Limited
  \item GUNKUL -- Gunkul Engineering Public Company Limited
  \item HANA -- Hana Microelectronics Public Company Limited
  \item INTUCH -- Intouch Holdings Public Company Limited
  \item IRPC -- IRPC Public Company Limited
  \item JMART -- Jaymart Group Holdings Public Company Limited
  \item MINT -- Minor International Public Company Limited
  \item PLANB -- Plan B Media Public Company Limited
  \item PTG -- PTG Energy Public Company Limited
  \item PTT -- PTT Public Company Limited
  \item QH -- Quality Houses Public Company Limited
  \item SINGER -- Singer Thailand Public Company Limited
  \item SPRC -- Star Petroleum Refining Public Company Limited
  \item TU -- Thai Union Group Public Company Limited
  \item VGI -- VGI Public Company Limited
  \item WHA -- WHA Corporation Public Company Limited
\end{itemize}

\end{document}